\documentclass{article}

\newif\ifdraft
\drafttrue

\usepackage{arxiv}
\usepackage[utf8]{inputenc} 
\usepackage[T1]{fontenc}    
\usepackage{hyperref}       
\usepackage{url}            
\usepackage{booktabs}       
\usepackage{amsfonts}       
\usepackage{nicefrac}       
\usepackage{microtype}      
\usepackage{lipsum}
\usepackage{amsmath}
\usepackage{balance}
\usepackage{booktabs} 
\usepackage{graphicx}
\usepackage{multirow}
\usepackage{rotating}
\usepackage{soul}
\usepackage{url}
\usepackage[utf8]{inputenc}
\usepackage[table]{xcolor}
\usepackage{subcaption}
\usepackage[export]{adjustbox}
\usepackage{stackengine}
\usepackage{amsfonts}
\usepackage{amssymb}
\usepackage{amsthm}
\usepackage{natbib}

\newcommand{\remove}[1]{\vspace{0ex}}

\newcommand{\jadci}{\texttt{JaDCI}}
\newcommand{\pydci}{\texttt{PyDCI}}

\newcolumntype{.}{D{.}{.}{-1}}
\newcolumntype{M}[1]{>{\centering\arraybackslash}m{#1}}
\newcolumntype{N}{@{}m{0pt}@{}}
\newcolumntype{C}[1]{>{\centering\let\newline\\\arraybackslash\hspace{0pt}}m{#1}}
\newcolumntype{L}{>{\raggedleft\arraybackslash}m{1cm}}

\title{Revisiting \\ Distributional Correspondence Indexing: \\ A Python Reimplementation and New Experiments}

\newcommand{\shadow}{\cellcolor[HTML]{C0C0C0}}

\author{
  Alejandro Moreo, Andrea Esuli, Fabrizio Sebastiani \\
  Istituto di Scienza e Tecnologie dell'Informazione \\
  Consiglio Nazionale delle Ricerche \\
  56124 Pisa, Italy\\
  \texttt{\{firstname.lastname\}@isti.cnr.it} \\
}

\begin{document}
\maketitle

\begin{abstract}
  This paper introduces \pydci, a new implementation of
  \emph{Distributional Correspondence Indexing} (DCI) written in
  Python. DCI is a transfer learning method for cross-domain and
  cross-lingual text classification for which we had provided an
  implementation (here called \jadci) built on top of \texttt{JaTeCS},
  a Java framework for text classification. \pydci\ is a stand-alone
  version of DCI that exploits \texttt{scikit-learn} and the
  \texttt{SciPy} stack.
  We here report on new experiments that we have carried out in order
  to test \pydci, and in which we use as baselines new high-performing
  methods that have appeared after DCI was originally proposed.  These
  experiments show that, thanks to a few subtle ways in which we have
  improved DCI, \pydci\ outperforms both \jadci\ and the
  above-mentioned high-performing methods, and delivers the best known
  results on the two popular benchmarks on which we had tested DCI,
  i.e., MultiDomainSentiment (a.k.a.\ MDS -- for cross-domain
  adaptation) and Webis-CLS-10 (for cross-lingual adaptation). \pydci,
  together with the code allowing to replicate our experiments, is
  available at \url{https://github.com/AlexMoreo/pydci} .
\end{abstract}



\section{Introduction}

\noindent \emph{Distributional Correspondence Indexing} (DCI) is a
pivot-based feature-transfer domain adaptation method for cross-domain
and cross-lingual text classification. DCI was first described in
\citep{dciecir2015}, and later improved and extended in
\citep{Moreo:2016fg}; it was formerly implemented in Java as part of
the \texttt{JaTeCS} (\underline{Ja}va \underline{Te}xt
\underline{C}ategorization \underline{S}ystem) framework
\citep{Esuli:2017wd}, and this implementation (henceforth called
\jadci) was made publicly
available.\footnote{\url{https://github.com/jatecs/jatecs/blob/master/src/main/java/it/cnr/jatecs/representation/transfer/dci/DistributionalCorrespondeceIndexing.java}}\texttt{JaTeCS}
is a complex package, since it makes available many functionalities
for text analytics research. A drawback of \jadci\ is thus that, for
the researcher wishing to replicate the results of
\citep{Moreo:2016fg} or simply wishing to use \jadci, a substantive
effort in installing and properly configuring the entire
\texttt{JaTeCS} framework is thus needed.

In this paper we present \pydci, a new implementation of the DCI
method written in Python and built on top of the \texttt{SciPy} stack
and \texttt{scikit-learn} toolkit. Python has become the preferred
programming language for computer scientists. In the fields of machine
learning and data mining its use has also been promoted by the
appearance of Python-based environments such as \texttt{SciPy} and
\texttt{scikit-learn}, whose potential and ease of use have attracted
the interest of practitioners. Our reimplementation is thus in line
with these trends.

With respect to \jadci, \pydci\ introduces a few modifications in the
way DCI is implemented that, although subtle, bring about a
significant improvement in the effectiveness of the method.

The rest of this paper is structured as follows. In Section
\ref{sec:changes} we describe the main modifications to DCI that our
new implementation introduces. In Section \ref{sec:results} we report
on new experiments that we have run using \pydci, and show that,
thanks to the modifications above, \pydci\ delivers new
state-of-the-art results on two popular benchmark datasets, i.e.,
MultiDomainSentiment (hereafter MDS -- for cross-domain adaptation)
and Webis-CLS-10 (for cross-lingual adaptation). These results
represent a clear improvement over the ones originally obtained with
\jadci\ and presented in \cite{Moreo:2016fg}, and also over the ones
obtained by recent high-performing methods that have appeared
\cite{ganin2016domain,Li:2017kg,xu2017cross,zhou2016cross}, or that we
have become aware of \cite{Yang:2015nb}, after DCI was originally
proposed. Section \ref{sec:conclusion} concludes, hinting at future
developments.

We make \pydci\ publicly available via
GitHub.\footnote{\url{https://github.com/AlexMoreo/pydci}}


\section{Implementation Changes}\label{sec:changes}

\noindent For reasons of brevity we do not re-explain DCI from
scratch;
we refer the interested reader to \citep{Moreo-Fernandez:2018ss} for a
concise description, or to \citep{Moreo:2016fg} for the full-blown
presentation.

The main modifications that \pydci\ introduces with respect to \jadci\
are the following:

\begin{enumerate}

\item \label{item:standardization} \textbf{Document Standardization}:
  In DCI, feature vectors and document vectors (i.e., the vectors that
  represent the features and the vectors that represent the documents,
  respectively) are post-processed via L2-normalization. In
  \citep{Moreo:2016fg} we had witnessed improvements when applying
  standardization to the feature vectors (i.e., translating and
  scaling each dimension so that it is approximatelly normally
  distributed in $\mathcal{N}(0,1)$ -- see \citep[p.\
  144]{Moreo:2016fg}). In \pydci\ we give the user the option to apply
  standardization also to each dimension of the document vectors
  before training the classifier.
  All experiments we report in this paper are run with this option
  activated.

\item \label{item:optimization} \textbf{Classifier Optimization}: In
  \pydci\ we use \texttt{scikit-learn}'s implementation of linear SVMs
  (\texttt{LinearSVC}, which is in turn based on the
  \texttt{liblinear}
  package\footnote{\url{https://www.csie.ntu.edu.tw/~cjlin/liblinear/}}),
  instead of using Joachims' SVM$^{light}$
  package\footnote{\url{http://svmlight.joachims.org/}} as we had done
  in \jadci.  This allows us to leverage \texttt{scikit-learn}'s
  \texttt{GridSearchCV} utility in order to optimize SVM's $C$
  parameter (which determines the trade-off between training error and
  the margin) via grid search optimization, which allows us to
  effortlessly tune the classifier. In the new experiments using
  \pydci\ we let parameter $C$ range in $\{10^i\}_{-5\leq i\leq 5}$,
  while in the \jadci\ experiments we had simply relied on the default
  value that SVM$^{light}$ attributes to $C$.

\item \label{item:pivots} \textbf{Increase in the Number of Pivots}:
  We increase the number of pivots from 100 (the value we had used in
  \citep{Moreo:2016fg}) to 1,000 in the cross-domain experiments and
  to 450 in the cross-lingual experiments. This brings about a
  significant improvement in performance, that does not come at a
  significant cost in execution time (as instead had happened with the
  previous implementation).  We limit the number of pivots to 450 in
  the cross-lingual case (instead of 1,000) since in this case each
  pivot requires a translation to the target language\footnote{A word
  translator oracle is used to automate the translation work in the
  experiments, following the indications of
  \citep{Prettenhofer:2010ys} and the bilingual dictionaries they
  released.} which is assumed to have a cost; we thus set the number
  of pivots to 450 as was done in previous research (e.g., in
  \citep{Prettenhofer:2010ys}). We discuss below in more detail the
  impact on performance that the variation in the number of pivots
  has.

\end{enumerate}

\noindent We should also mention that \pydci\ relies on
\texttt{scikit-learn} (while \jadci\ relied on \texttt{JaTeCS}) for
many preprocessing-related aspects (e.g., term weighting), which also
may cause some (hard to track) differences in performance with respect
to \jadci.


\section{Experiments}
\label{sec:results}


\subsection{Effectiveness on Cross-Domain Classification and
Cross-Lingual Classification}
\label{sec:effectiveness}

\noindent In this section we report the results we have obtained in
rerunning with \pydci\ the same experiments we had run with \jadci,
and whose results had been reported in \citep{Moreo:2016fg}.  The
datasets we use are arguably the most popular benchmarks in the domain
adaptation literature, i.e., MDS \citep{Blitzer:2007gf} for
cross-domain
adaptation\footnote{\url{http://www.cs.jhu.edu/~mdredze/datasets/sentiment/}}
and Webis-CLS-10 \citep{Prettenhofer:2010ys} for cross-lingual
adaptation\footnote{\url{https://www.uni-weimar.de/en/media/chairs/computer-science-department/webis/data/corpus-webis-cls-10/}}.
A complete description of the datasets and the standard experimental
protocol followed in each case can be found either in the original
publications describing the datasets
\citep{Blitzer:2007gf,Prettenhofer:2010ys} or in \citep{Moreo:2016fg}.

Tables \ref{tab:mds} and \ref{tab:webis} show the values of
classification accuracy (i.e., the fraction of correctly classified
documents) we obtain for cross-domain and cross-lingual classification
experiments, respectively.  We focus on Linear and Cosine (columns
9-10), two parameter-free probabilistic and kernel-based
\emph{distributional correspondence functions} (DCFs) investigated in
\citep{Moreo:2016fg}. For each such DCF we show a direct comparison
against the values we had obtained with \jadci\ (Columns 7-8).  We
also report two baselines:

\begin{itemize}

\item \textsc{Lower} (Column 3), a classifier that directly trains on
  the ``source'' training examples and tests on the ``target''
  unlabeled examples without performing any sort of adaptation at
  all. Such a classifier should thus act as a lower bound for any
  reasonable adaptation endeavour.

\item \textsc{Upper} (Column 4), a classifier that trains on the
  ``target'' training examples and tests on the ``target'' unlabeled
  examples without performing any sort of adaptation at
  all.\footnote{In MDS there is only one labelled set available for
  each domain (see \citep{Blitzer:2007gf}). In this case we report the
  accuracy of a 5-fold cross-validation on the test set.} Such a
  classifier should thus act as an upper bound for any reasonable
  adaptation endeavour.

\end{itemize}

\noindent The baselines use exactly the same learner we use for
\pydci\ (\texttt{LinearSVC} with the $C$ parameter optimized via grid
search). For each (problem, dataset) pair we also report the accuracy
obtained by what, to the best of our knowledge, is today the
best-performing known method on this (problem, dataset) pair (Column 5
-- labelled as ``SOTA'', which stands for ``\underline{S}tate
\underline{O}f \underline{T}he \underline{A}rt'' -- reports the name
of the method and Column 6 reports the accuracy score, taken from the
original paper).  Boldface indicates the best score for each (problem,
dataset) pair; shadowed cells indicate the \pydci\ scores that
outperform the best-known results.

Note that, aside from SDA \citep{glorot2011domain}, all the baselines
in the ``SOTA'' column had not been used as baselines in our original
work on DCI; the reason is that these methods were published after DCI
appeared in print
\citep{ganin2016domain,Li:2017kg,xu2017cross,zhou2016cross}, or that
we were unaware of them \citep{Yang:2015nb}.

\begin{table*}[t]
  \centering
  \resizebox{\textwidth}{!}{%
  \begin{tabular}{|c|c||c|c||r|c||c|c||c|c|}
    \hline
    \multicolumn{2}{|c||}{Task} & \multicolumn{2}{c||}{Baselines} & \multicolumn{2}{c||}{SOTA} & \multicolumn{2}{c||}{\jadci} & \multicolumn{2}{c|}{\pydci} \\
    \hline
    Source & Target & Lower & Upper & method & score & Linear & Cosine & Linear & Cosine \\\hline\hline
                                & DVD & 0.807 & 0.850 & SDA \citep{glorot2011domain} & \textbf{0.844} & 0.808 & 0.817 & 0.803 & 0.823 \\
                                & Electronics & 0.734 & 0.871 & AMN \citep{Li:2017kg} & 0.808 & 0.810 & 0.822 & \cellcolor[HTML]{C0C0C0}\textbf{0.837} & \cellcolor[HTML]{C0C0C0}\textbf{0.837} \\
    \multirow{-3}{*}{Books} & Kitchen & 0.774 & 0.907 & DANN \citep{ganin2016domain} & 0.843 & 0.834 & 0.835 & \cellcolor[HTML]{C0C0C0}\textbf{0.851} & \cellcolor[HTML]{C0C0C0}0.843 \\\hline
                                & Books & 0.790 & 0.839 & DANN \citep{ganin2016domain} & 0.825 & 0.825 & 0.824 & \cellcolor[HTML]{C0C0C0}0.832 & \cellcolor[HTML]{C0C0C0}\textbf{0.835} \\
                                & Electronics & 0.757 & 0.871 & CDFL \citep{Yang:2015nb} & 0.809 & 0.822 & 0.824 & \cellcolor[HTML]{C0C0C0}0.839 & \cellcolor[HTML]{C0C0C0}\textbf{0.855} \\
    \multirow{-3}{*}{DVD} & Kitchen & 0.778 & 0.907 & DANN \citep{ganin2016domain} & 0.849 & 0.858 & \textbf{0.864} & \cellcolor[HTML]{C0C0C0}0.853 & \cellcolor[HTML]{C0C0C0}0.856 \\\hline
                                & Books & 0.716 & 0.839 & AMN \citep{Li:2017kg} & 0.780 & 0.766 & 0.764 & \cellcolor[HTML]{C0C0C0}0.796 & \cellcolor[HTML]{C0C0C0}\textbf{0.800} \\
                                & DVD & 0.745 & 0.850 & DANN \citep{ganin2016domain} & 0.781 & 0.768 & 0.774 & \cellcolor[HTML]{C0C0C0}0.787 & \cellcolor[HTML]{C0C0C0}\textbf{0.801} \\
    \multirow{-3}{*}{Electronics} & Kitchen & 0.859 & 0.907 & SDA \citep{glorot2011domain} & \textbf{0.902} & 0.864 & 0.868 & 0.871 & 0.878 \\\hline
                                & Books & 0.737 & 0.839 & AMN \citep{Li:2017kg} & 0.793 & 0.783 & 0.790 & 0.779 & \cellcolor[HTML]{C0C0C0}\textbf{0.807} \\
                                & DVD & 0.746 & 0.850 & CDFL \citep{Yang:2015nb} & \textbf{0.876} & 0.788 & 0.799 & 0.795 & 0.806 \\
    \multirow{-3}{*}{Kitchen} & Electronics & 0.840 & 0.871 & SDA \citep{glorot2011domain} & \textbf{0.872} & 0.855 & 0.858 & 0.853 & 0.860 \\\hline\hline
                                & Average & 0.773 & 0.867 & AMN \citep{Li:2017kg} & 0.814 & 0.815 & 0.820 & \cellcolor[HTML]{C0C0C0}0.825 & \cellcolor[HTML]{C0C0C0}\textbf{0.833}\\\hline
  \end{tabular}
  }
  \caption{Cross-domain classification on the MDS dataset.}
  \label{tab:mds}
\end{table*}

\begin{table*}[t]
  \centering
  \resizebox{\textwidth}{!}{%
  \begin{tabular}{|c|c||c|c||r|c||c|c||c|c|}
    \hline
    \multicolumn{2}{|c||}{Task} & \multicolumn{2}{c||}{Baselines} & \multicolumn{2}{c||}{SOTA} & \multicolumn{2}{c||}{\jadci} & \multicolumn{2}{c|}{\pydci} \\
    \hline
    \begin{tabular}[c]{@{}c@{}}Target \\ Language\end{tabular} & Domain  & Lower & Upper & method  & score & Linear & Cosine & Linear & Cosine \\\hline\hline
    \multirow{3}{*}{German} & Books & 0.523 & 0.863 & BiDRL \citep{zhou2016cross} & 0.841 & 0.798 & 0.827 & \shadow 0.846 & \shadow \textbf{0.850} \\
                                & DVD & 0.562 & 0.837 & BiDRL \citep{zhou2016cross} & \textbf{0.841} & 0.826 & 0.822 & \shadow \textbf{0.841} & 0.837 \\
                                & Music & 0.558 & 0.849 & BiDRL \citep{zhou2016cross} & 0.847 & 0.844 & 0.856 & \shadow \textbf{0.865} & \shadow 0.852 \\\hline
    \multirow{3}{*}{French} & Books & 0.558 & 0.844 & BiDRL \citep{zhou2016cross} & \textbf{0.844} & 0.746 & 0.842 & 0.834 & 0.816 \\
                                & DVD & 0.537 & 0.843 & BiDRL \citep{zhou2016cross} & 0.836 & 0.823 & 0.827 & 0.835 & \shadow \textbf{0.851} \\
                                & Music & 0.566 & 0.876 & CLDFA \citep{xu2017cross} & 0.833 & 0.816 & \textbf{0.844} & 0.824 & \shadow 0.842 \\\hline
    \multirow{3}{*}{Japanese} & Books & 0.498 & 0.802 & CLDFA \citep{xu2017cross} & 0.774 & 0.779 & 0.758 & \shadow \textbf{0.796} & \shadow 0.790 \\
                                & DVD & 0.500 & 0.814 & CLDFA \citep{xu2017cross} & 0.805 & 0.822 & 0.801 & \shadow \textbf{0.830} & \shadow 0.802 \\
                                & Music & 0.509 & 0.834 & BiDRL \citep{zhou2016cross} & 0.788 & 0.826 & \textbf{0.839} & \shadow 0.811 & \shadow 0.838 \\\hline\hline
                                & Average & 0.534 & 0.840 & BiDRL \citep{zhou2016cross} & 0.813 & 0.809 & 0.824 & \shadow \textbf{0.831} & \shadow \textbf{0.831} \\\hline
  \end{tabular}
  }
  \caption{Cross-lingual classification on the Webis-CLS-10 dataset.}
  \label{tab:webis}
\end{table*}

\pydci\ outperforms \jadci\ in most cases, and outperforms also the
best-performing method in the literature, which is not always the same
for each (problem,dataset) pair, with very few exceptions.  \pydci\
obtains 7 out of 13 best results on MDS (including best averaged
accuracy) when equipped with the Cosine DCF, and 5 out of 10 best
results in Webis-CLS-10 when using the Linear DCF (including best
averaged accuracy).  In agreement with with \citep{Moreo:2016fg},
Cosine proved the best performing DCF, yielding the best results
overall and surpassing the best accuracy obtained by any other method
in 17 cases out of 23 (across the two datasets, and also including the
average results).  With respect to the previously best-performing
system, \pydci(Cosine) brings about a reduction in error of +10.2\% on
MDS and +9.6\% on Webis-CLS-10.

On the very same (problem,dataset) pairs we have also run experiments
in order to evaluate the impact of modifications
\ref{item:standardization} (Document Standardization) and
\ref{item:optimization} (Classifier Optimization) mentioned in Section
\ref{sec:changes}. Concerning document standardization, we have rerun
all the \pydci\ experiments described in Tables \ref{tab:mds} and
\ref{tab:webis} without applying document standardization. The results
are reported in the first two rows of Table \ref{tab:modifications},
and indicate, on average, a relative improvement in accuracy of +0.2\%
on MDS and +8.6\% on Webis-CLS-10; document standardization thus
appears to be clearly beneficial. Concerning classifier optimization,
we have rerun all the \pydci\ experiments described in Tables
\ref{tab:mds} and \ref{tab:webis} without applying classifier
optimization. The results are reported in the last two rows of Table
\ref{tab:modifications}, and indicate, on average, a relative
improvement in accuracy of +9.7\% on MDS and +11.8\% on Webis-CLS-10;
also classifier optimization is thus (unsurprisingly) clearly
beneficial.\footnote{Note that the results obtained by \pydci\ without
document standardization and classifier optimization are different
from the ones obtained by \jadci, the main reason being that
SVM$^{light}$ and \texttt{LinearSVC} choose different default
parameters for their SVM learner.}

\begin{table}[t]
  \begin{center}
    \begin{tabular}{|r|c||cc|cc|}
      \hline
      & & \multicolumn{2}{c|}{MDS} & \multicolumn{2}{c|}{Webis-CLS-10} \\
      \hline\hline
      \multirow{2}{*}{Document Standardization} & with    & 0.833 & (+0.2\%) & 0.831 & (+8.6\%) \\
      & without & 0.831 &          & 0.757 &    \\
      \hline
      \multirow{2}{*}{Classifier Optimization} & with    & 0.833 & (+9.7\%) & 0.831 & (+11.8\%) \\
      & without & 0.767 &          & 0.743 &    \\
      \hline
    \end{tabular}
  \end{center}
  \caption{\label{tab:modifications}Average accuracy obtained using
  \pydci\ (using the Cosine DCF) with or without document
  standardization and classifier optimization; percentages indicate
  relative improvement of the ``with'' configuration with respect to
  the corresponding ``without'' configuration.}
\end{table}%


\subsection{Effectiveness on Cross-Domain Cross-Lingual
Classification}
\label{sec:CDCL}

\noindent Table \ref{tab:webis2} reports classification accuracy
values obtained in the domain adaptation setting proposed in
\citep{Moreo:2016fg}, in which both domain \emph{and} language differ
between the source and target (i.e., when the classification task is
simultaneously cross-domain and cross-lingual). In Table
\ref{tab:webis2} we include the results we had obtained in
\citep{Moreo:2016fg} for the Cross-Lingual Structural Correspondence
Learning (SCL) method \citep{Prettenhofer:2010ys} (which we use here
as a baseline), using its authors'
code\footnote{\url{https://github.com/pprett/nut}} (see
\citep{prettenhofer2011cross}). The reason why we use SCL as a
baseline is that, although newer approaches have been tested in this
setting, none of them, to the best of our knowledge, has outperformed
SCL so far.

The results in Table \ref{tab:webis2} confirm the superiority of
\pydci\ over \jadci. In this case, though, the differences in
performance between the ``Cosine'' counterparts is less pronounced.
Between the \pydci\ variants, Linear performs slightly better than
Cosine.

\begin{table*}[ht!]
  \centering
  \begin{tabular}{|c|c||c|c|c||c|c||c|c|}
    \hline
    \multicolumn{2}{|c||}{Task} & \multicolumn{3}{c||}{Baselines} & \multicolumn{2}{c||}{\jadci} & \multicolumn{2}{c|}{\pydci}       \\ \hline
    Source     & Target      & Lower         & Upper         & SCL 
    & Linear   & Cosine          & Linear         & Cosine         \\\hline\hline
    EB         & GD          & 0.495         & 0.837         & 0.784 & 0.790    & 0.827           & \textbf{0.834} & 0.812          \\
    EB         & GM          & 0.525         & 0.849         & 0.811 & 0.786    & 0.843           & \textbf{0.861} & 0.853          \\
    EB         & FD          & 0.533         & 0.843         & 0.780 & 0.810    & 0.823           & \textbf{0.841} & 0.832          \\
    EB         & FM          & 0.533         & 0.876         & 0.762 & 0.822    & 0.833           & 0.834          & \textbf{0.854} \\
    EB         & JD          & 0.491         & 0.814         & 0.742 & 0.813    & 0.805           & \textbf{0.830} & 0.816          \\
    EB         & JM          & 0.480         & 0.834         & 0.742 & 0.826    & \textbf{0.831}  & 0.826          & 0.826          \\ \hline
    ED         & GB          & 0.568         & 0.863         & 0.823 & 0.823    & 0.824           & \textbf{0.855} & 0.808          \\
    ED         & GM          & 0.561         & 0.849         & 0.824 & 0.844    & 0.816           & 0.846          & \textbf{0.851} \\
    ED         & FB          & 0.511         & 0.844         & 0.790 & 0.744    & \textbf{0.848}  & 0.839          & 0.828          \\
    ED         & FM          & 0.540         & 0.876         & 0.757 & 0.836    & 0.847           & 0.817          & \textbf{0.866} \\
    ED         & JB          & 0.505         & 0.802         & 0.725 & 0.738    & 0.761           & \textbf{0.775} & 0.769          \\
    ED         & JM          & 0.484         & 0.834         & 0.776 & 0.817    & 0.816           & \textbf{0.825} & 0.824          \\ \hline
    EM         & GB          & 0.536         & 0.863         & 0.825 & 0.791    & 0.812           & 0.811          & \textbf{0.846} \\
    EM         & GD          & 0.539         & 0.837         & 0.792 & 0.778    & \textbf{0.834}  & 0.793          & \textbf{0.834} \\
    EM         & FB          & 0.523         & 0.844         & 0.784 & 0.810    & \textbf{0.845}  & 0.816          & 0.812          \\
    EM         & FD          & 0.559         & 0.843         & 0.745 & 0.798    & 0.841           & 0.817          & \textbf{0.847} \\
    EM         & JB          & 0.503         & 0.802         & 0.708 & 0.711    & 0.721           & \textbf{0.785} & 0.746          \\
    EM         & JD          & 0.511         & 0.814         & 0.756 & 0.792    & 0.790           & \textbf{0.823} & 0.801          \\ \hline \hline
                                & Average     & 0.522         & 0.840         & 0.774 & 0.796    & 0.818           & \textbf{0.824} & 0.823  \\\hline     
  \end{tabular}
  \caption{Cross-domain cross-lingual classification on the
  Webis-CLS-10 dataset. In the first two columns, for conciseness we
  use the following notation: E (English), G (German), F (French), and
  J (Japanese) for languages; and B (Books), D (DVD), and M (Music)
  for domains. E.g., GB stands for German-Books.}
  \label{tab:webis2}
\end{table*}


\subsection{Statistical Significance}
\label{sec:stat}

\noindent We have subjected our experiments to thorough statistical
significance testing, by running a two-tailed t-test on paired
examples across all runs (cross-domain and/or cross-lingual). The test
reveals that the \pydci\ versions of Linear and Cosine outperform, in
a statistically significant sense, the corresponding \jadci\ versions
(at a confidence level of $\alpha=0.005$).


\subsection{Efficiency}
\label{sec:efficiency}

\begin{figure*}[ht!]
  \centering \includegraphics[width=0.75\textwidth]{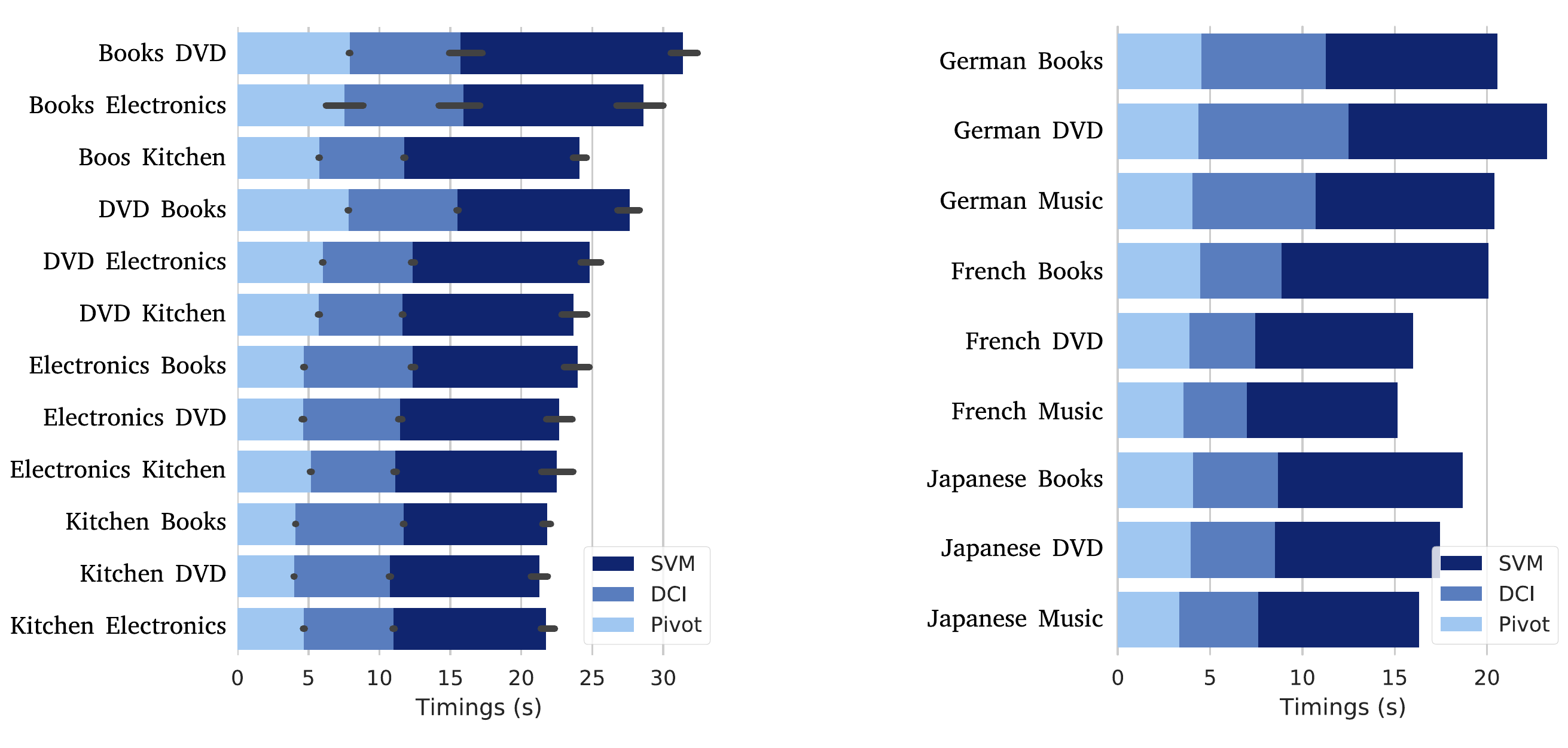}
  \caption{Computation times for the MDS (left) and Webis-CLS-10
  (right) datasets. Values include the time required for pivot
  selection (Pivot), DCI projection (DCI), and SVM training and
  optimization (SVM). The time for 
  preprocessing the documents is not included. Results for MDS are
  averages across 5 folds (see \citep{Blitzer:2007gf} for further
  details).}
  \label{fig:timings}
\end{figure*}

\noindent One important aspect of DCI in general, and of \pydci\ in
particular, is its efficiency. Figure \ref{fig:timings} reports the
computation times we have recorded\footnote{The experiments were run
on a machine equipped with a 8-core processor AMD FX-8350 at 4GHz with
32 GB of RAM under Ubuntu 16.04 (LTS).} in order to measure the
efficiency of \pydci.  While the best-performing methods from the
literature rely on computationally expensive optimizations (most of
them are deep-learning-based), none of the experiments we have
presented so far required more than 35 seconds to run.

\subsection{Effectiveness vs.\ Efficiency Trade-off}
\label{sec:effvseff}

\begin{figure*}[ht!]
  \centering
  \includegraphics[width=0.9\textwidth]{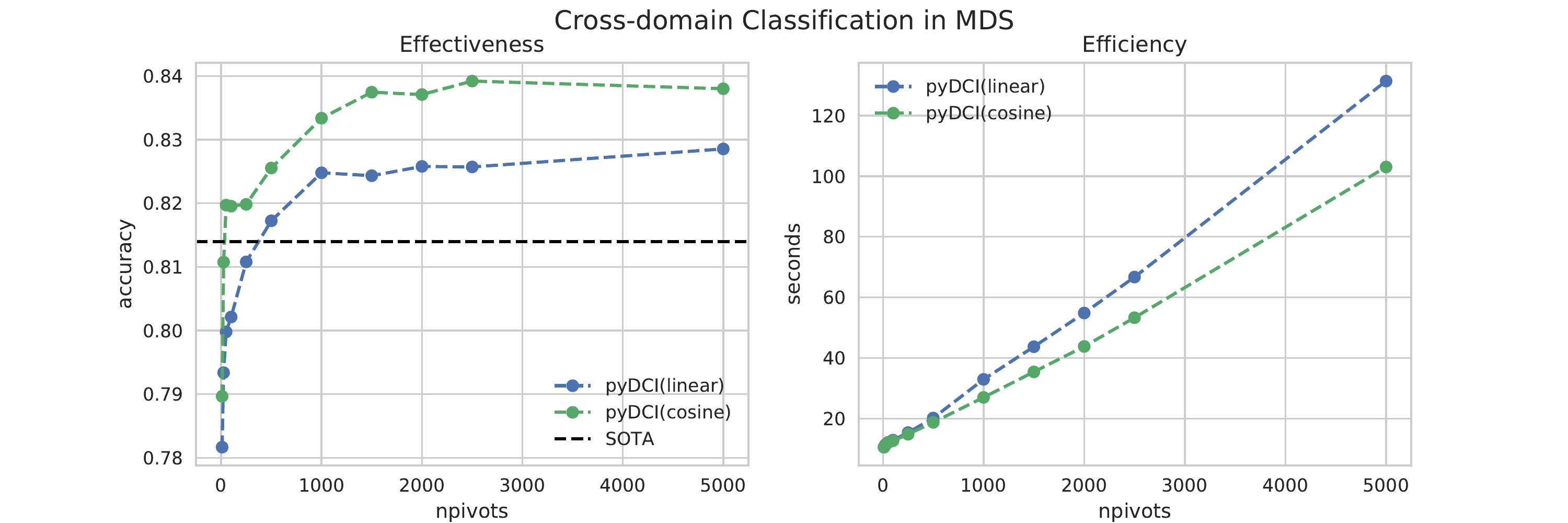}
  \includegraphics[width=0.9\textwidth]{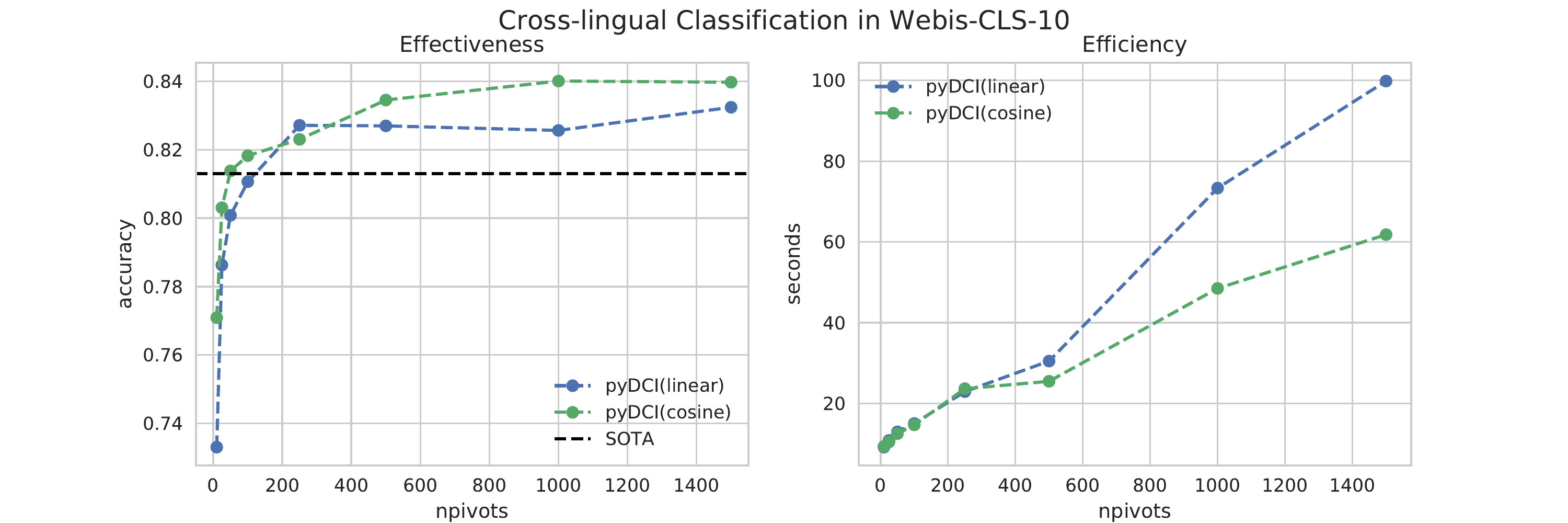}
  \caption{Effectiveness (left) and efficiency (right) of \pydci\ as a
  function of the number of pivots, for cross-domain (top) and
  cross-lingual (bottom) sentiment classification. }
  \label{fig:varying}
\end{figure*}

\noindent In this section we analyse the trade-off between
effectiveness (in terms of classification accuracy) and time
efficiency (in terms of seconds).  In this experiment, we vary the
number of pivots in the range
$[10,25,50,100,250,500,1000,1500,2000,2500,5000]$.  For the
Webis-CLS-10 we bound this range to $1500$ pivots since, for some
tasks it was impossible to extract more than $1500$ pivots.  Figure
\ref{fig:varying} shows the average accuracy (left) and computation
times for MDS and Webis-CLS-10.

As $npivots$ increases, \pydci\ surpasses the best average accuracy
reported for any other method in both datasets. In particular, and in
accordance with \citep{Moreo:2016fg}, \pydci\ equipped with the Cosine
DCF does so with only 100 pivots.  In this case, and in contrast with
\jadci, classification accuracy increases noticeably when more pivots
are taken into account; this might be a side effect of the
modifications discussed in Section \ref{sec:changes}.  In any case,
the method seems to reach a plateau for higher values of $npivots$,
allowing the Cosine variant to reach new peaks of classification
accuracy of 0.839 (when $npivots=2500$) in MDS, and 0.840 (when
$npivots=1000$) in Webis-CLS-10.  Regarding the efficiency of the
method, \pydci\ exhibits a quasi-linear trend in time complexity,
e.g., when the number of pivots is doubled, the execution time is
roughly doubled too.


\section{Conclusions}
\label{sec:conclusion}

\noindent We have presented \pydci, a (Python-based) revision of our
previous (Java-based) implementation of DCI.  This new implementation
incorporates changes that, although subtle, nonetheless allow the
method to deliver improved results that outperform the currently known
best-performing methods.  The efficiency tests we have carried out
speak clearly about the efficiency of \pydci, which requires roughly
half a minute to undertake any of the domain adaptation tasks in our
experiments.

In a preliminary study DCI was also tested in transductive scenarios
\citep{Moreo:2016pa}.  \pydci\ does not support transductive
classification; this is something we plan to address in the near
future.


\bibliographystyle{plainnat}
\bibliography{references}
\end{document}